\documentclass[10pt,twocolumn,letterpaper]{article}

\usepackage{iccv}
\usepackage{times}
\usepackage{epsfig}
\usepackage{graphicx}
\usepackage{amsmath}
\usepackage{amssymb}
\usepackage{stmaryrd}
\usepackage[accsupp]{axessibility}

\usepackage{comment}
\usepackage{xfrac}
\newcommand{\vn}[1]{\ensuremath{{\mathbf #1}}}
\newcommand{\PT}[2]{\ensuremath{{\mathcal P}_{#1 \shortleftarrow #2}}}
\newcommand{\eval}[2]{\ensuremath{\left. #1 \right\rvert_{#2}}}
\newcommand{\N}{\ensuremath{{\mathcal N}}}
\newcommand{\abs}[1]{\ensuremath{\left \vert #1 \right \vert}}

\usepackage[pagebackref=true,breaklinks=true,letterpaper=true,colorlinks,bookmarks=false]{hyperref}

\iccvfinalcopy 

\def\iccvPaperID{}

\ificcvfinal\fi

\begin{document}

%%%%%%%%% TITLE
\title{Field Convolutions for Surface CNNs}

\author{Thomas W. Mitchel\\
\normalsize Johns Hopkins University\\
{\tt\small tmitchel@jhu.edu}
\and
Vladimir G. Kim\\
\normalsize Adobe Research\\
{\tt\small 
vokim@adobe.com}
\and
Michael Kazhdan\\
\normalsize Johns Hopkins University\\
{\tt\small 
misha@cs.jhu.edu}
}

\maketitle

\ificcvfinal\thispagestyle{empty}\fi

%%%%%%%%% ABSTRACT
\begin{abstract}
We present a novel surface convolution operator acting on vector fields that is based on a simple observation: instead of combining neighboring features with respect to a single coordinate parameterization defined at a given point, we have every neighbor describe the position of the point within its own coordinate frame.  This formulation combines intrinsic spatial convolution with parallel transport in a scattering operation while placing no constraints on the filters themselves, providing a definition of convolution that commutes with the action of isometries, has increased descriptive potential, and is robust to noise and other nuisance factors. The result is a rich notion of convolution which we call field convolution, well-suited for CNNs on surfaces. Field convolutions are flexible, straight-forward to incorporate into surface learning frameworks, and their highly discriminating nature has cascading effects throughout the learning pipeline. Using simple networks constructed from residual field convolution blocks, we achieve state-of-the-art results on standard benchmarks in fundamental geometry processing tasks, such as shape classification, segmentation, correspondence, and sparse matching. 

\end{abstract}

%%%%%%%%% Intro
\section{Introduction}
\label{s:intro}
The advent of deep learning in imaging, vision, and graphics has coincided with the development of numerous techniques for the analysis and processing of curved surfaces based on convolutional neural networks (\textbf{CNNs}). The challenge in reproducing the success of CNNs on surfaces is that classical notions of convolution and correlation in Euclidean spaces  cannot simply be transposed onto curved domains. Unlike images, points on a surface have no canonical orientation, without which
simple operations fundamental to the spatial propagation of information, such as moving dot products,
cannot be computed in a repeatable manner. 

Geometric deep learning is a young field, and many successful methods can be broadly categorized in relation to two emerging paradigms characterized by specific approaches to convolution: \textit{diffusive} propagation and \textit{equivariant} propagation. Diffusive approaches closely intertwine convolution operations with heat diffusion on manifolds wherein filters represented by anisotropic heat kernels or Gaussians are used to propagate scalar features \cite{masci2015geodesic,  boscaini2016anisotropic, boscaini2016learning, monti2017geometric, li2020shape, sharp2020diffusion}. In contrast, equivariant convolutions distribute vector or tensor features that transform with local coordinate systems \cite{poulenard2018multi, thomas2018tensor, schonsheck2018parallel, cohen2019gauge, de2020gauge, wiersma2020cnns, surfacecnn}.

Critically, virtually all state-of-the-art approaches sacrifice filter descriptiveness
to define a notion of convolution that does not depend on the choice of local coordinate frames. Gaussian filters can facilitate efficient evaluations in the spectral domain but are individually undiscriminating.
Extensions to anisotropic filtering mitigate these limitations by extending the class of filters that can be used. However, this requires defining a frame field over the surface -- itself a hard problem.

\begin{figure}[!t] 
\centering
\includegraphics[width=\columnwidth]{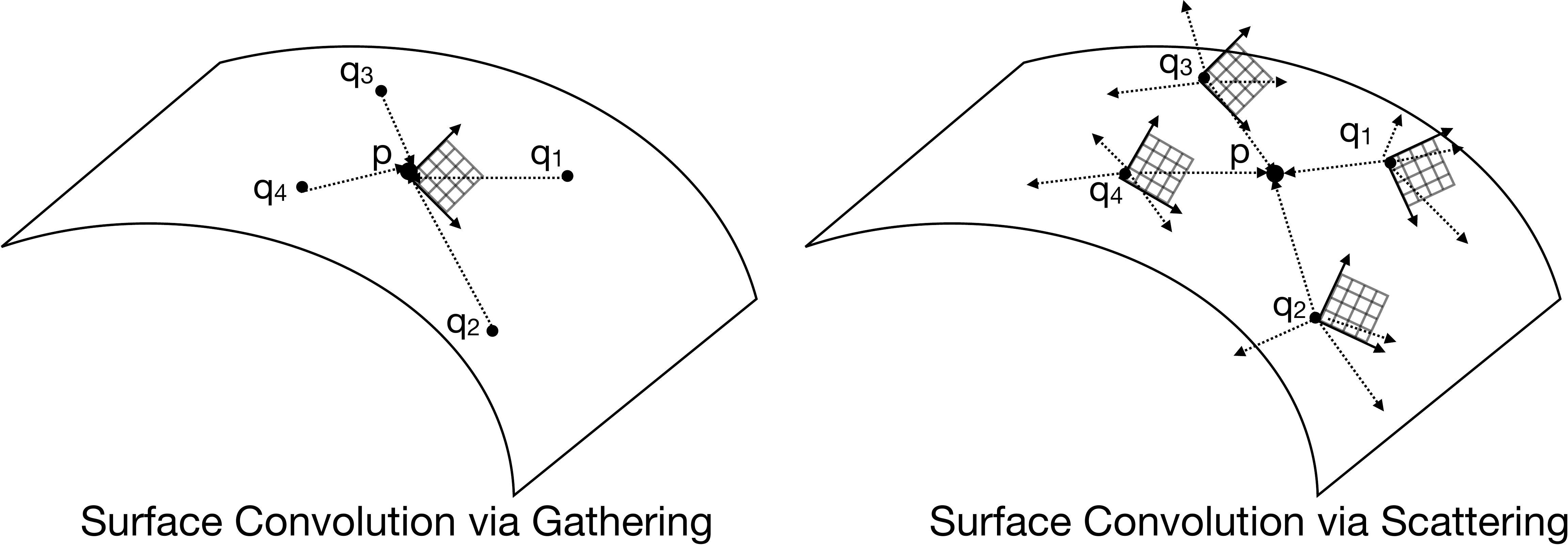}
\caption{Prior approaches define patch-based convolution operators as \textit{gathering} operations (left), which are sensitive to noise or disruptions in the local coordinate system. Field convolution is a \textit{scattering} operation and is robust under perturbations as it does not rely on a single coordinate system to aggregate features. \label{scatter_vs_gather}}
\end{figure}

Equivariant approaches have the potential to provide expressive notions of convolution on surfaces due to the encoding of geometric information in the transport of tangent vector features. However, equivariance of the response is almost universally achieved by placing constraints on the filters themselves \cite{poulenard2018multi, Poulenard2O19, Cohen2019, de2020gauge, curvanet2020, wiersma2020cnns}, limiting descriptiveness and necessitating complex architectures to support the algebraic relationships between kernels. Furthermore, these regimes formulate spatial propagation as a \textit{gathering} operation, analogous to correlations on Euclidean domains; features are weighted based on their position relative to a coordinate frame defined at a single point (Figure~\ref{scatter_vs_gather}, left), making them sensitive to inconsistencies or disruptions in local parameterizations .

In this paper we present a novel convolution operator acting on vector fields.
Our method defines the value of the convolution at a point $p$ using a simple observation: instead of combining neighboring features by parameterizing each neighbor $q_i$ with respect to a coordinate frame defined at $p$, each neighbor $q_i$ parametrizes $p$ within its own coordinate frames (Figure~\ref{scatter_vs_gather}, right).
This formulation combines intrinsic spatial weighting with parallel transport \textit{while placing no constraints on the filters themselves}, providing a definition of convolution that commutes with the action of isometries and has increased descriptive potential. In addition, as a \textit{scattering} operation, it is less sensitive to noise and other nuisance factors as it does not rely on a single coordinate system about each point to aggregate features. The result is a rich notion of convolution which we call \textit{field convolution} (\textbf{FC}), well-suited for CNNs on surfaces.

Field convolutions are flexible and straight-forward to incorporate into surface learning frameworks. Their highly discriminating nature has cascading effects throughout the learning pipeline, allowing us to achieve state-of-the-art results on standard benchmarks in applications including shape classification, segmentation, correspondence, and sparse matching.  All code and evaluations are publicly available at \href{https://github.com/twmitchel/FieldConv}{github.com/twmitchel/FieldConv}.

\section{Related Work}
\label{s:related}
The field of geometric deep learning has grown extensively since its inception half a decade ago. Here, we only review the techniques most closely related to ours -- those designed specifically for the analysis of 3D shapes. Generally speaking, these methods exist on a spectrum  between extrinsic and intrinsic techniques, with the former performing signal processing using the embedding of the surface in 3D and the latter only using the Riemannian structure.

\textit{Point-based} methods offer a purely extrinsic framework for applying deep learning to 3D shapes by representing them in terms of point clouds. A majority of these approaches can trace their lineage to the influential PointNet \cite{qi2017pointnet} and PointNet$++$ architectures \cite{qi2017pointnet++} and recent approaches such as DGCNN \cite{wang2019dynamic}, PCNN \cite{atzmon2018point}, KPCNN \cite{thomas2019kpconv}, TFN \cite{thomas2018tensor}, QEC \cite{zhao2020quaternion} and SPHNet \cite{Poulenard2O19} have sought to extend the framework by incorporating connectivity information, dynamic filter parameterizations, and equivariance to rigid transformations.  Convolution is typically expressed by applying
radially isotropic filters over local 3D neighborhoods and aggregating the results with the maximum or summation operations. This approach offers a simple foundation for extremely flexible and noise-robust networks, though at the expense of descriptive potential. More generally, these methods tend to struggle in the presence of non-rigid isometric deformations, making them less effective in scenarios like deformable shape matching \cite{donati2020deep, groueix20183d, sharp2020diffusion}.

\textit{Representational} approaches sit between extrinsic and intrinsic techniques. These methods exploit the data's underlying connectivity to form convolutional operators, often making use of well-developed techniques for graph-based learning on irregular structures \cite{cheby2016, Yi_2017_CVPR, verma2018feastnet, fey2018splinecnn, levie2018cayleynets, gong2019spiralnet++, Cohen2019, lahav2020meshwalker}. In particular, convolutions are performed using filters defined relative to the explicit graph structure as functions on edges or vertices, often with only immediate local support such as the surrounding one-ring or half-edge. A particularly notable example is MeshCNN \cite{hanocka2019meshcnn}, which specifically leverages the ubiquitous representation of surfaces as triangle meshes to construct a similarity-invariant convolution operator propagating edge-based features. 
While this enables graph-based convolutions to better handle non-rigid deformations compared to point-based approaches, it also makes them sensitive to changes in connectivity. 

One approach to defining {\em intrinsic} convolution has been to parametrize the surface over a simple domain such as the the sphere \cite{haim2019surface},  torus \cite{maron2017convolutional}, or plane \cite{sinha2016geoim} where standard CNNs can be applied. However, such parameterizations depend on the genus and often exhibit significant distortion.

A second class of approaches has been to define intrinsic convolution over the Riemannian manifold, and can generally be classified in relation to two emerging paradigms: \textit{diffusive} convolutions and \textit{equivariant} convolutions. In the former, convolution operations are closely related to heat diffusion on surfaces wherein heat (e.g. Gaussian) kernels are used to propagate scalar features. While early diffusive approaches including GCNN \cite{masci2015geodesic}, ADD \cite{boscaini2016learning}, ACNN \cite{boscaini2016anisotropic} and  MoNet \cite{monti2017geometric} perform convolutions over local patches, recent state-of-the-art networks ACSCNN \cite{li2020shape} and DiffusionNet \cite{sharp2020diffusion} represent convolution in the spectral domain. 
Despite their success in a variety of scenarios, most notably in dense shape correspondence \cite{groueix20183d, donati2020deep, li2020shape, sharp2020diffusion}, these methods face an intractable problem: radially symmetric filters are individually undiscriminating and diffusive frameworks are not naturally suited to handle the orientation ambiguity problem introduced by the use of more descriptive, anisotropic kernels. To compensate, these methods supplement convolutions with basic orientation-aware operations on tangent vector features \cite{sharp2020diffusion} in addition to employing various strategies that are either fragile, such as aligning kernels along the directions of principal curvature \cite{boscaini2016anisotropic, monti2017geometric}, or discarding information by pooling over samplings of orientations or by specifying directions of maximum activation\cite{masci2015geodesic, li2020shape}.

Recently, several techniques have been introduced for \textit{equivariant} surface convolutions such as MDGCNN \cite{poulenard2018multi}, GCN \cite{cohen2019gauge, de2020gauge} and HSN \cite{wiersma2020cnns}. In contrast to diffusive approaches, equivariant convolutions are designed specifically to address the rotation ambiguity problem by propagating tangent vector features that transform with local coordinate systems. 
To make the convolution independent of the choice of local coordinate frame, most existing methods strongly constrain the class of filters that can be used  \cite{poulenard2018multi, Poulenard2O19, Cohen2019, de2020gauge, curvanet2020, wiersma2020cnns}. An exception to this is PFCNN \cite{surfacecnn} which also discards information by pooling over multiple kernel orientations.  Often, these parameterizations are so restrictive that they necessitate complex network architectures to be effective: even the state-of-the-art HSN \cite{wiersma2020cnns} is formulated as a multi-stream U-Net with various pooling operations. Furthermore, in moving from radially isotropic to anisotropic filters, prior equivariant regimes universally formulate spatial propagation as a \textit{gathering} operation wherein all features in the local surface are weighted based on their position in a \textit{single} coordinate system. While this approach may seem natural as it is analogous to correlation on Euclidean domains, a feature's dependence on a single local parameterization increases sensitivity to noise.% and changes in sampling density. 

\section{Method Overview}
\label{s:overview}
Field convolutions are closely related to an operation called \textit{extended convolution}, which allows a filter to adaptively transform as it travels over a Euclidean domain or manifold \cite{mitchel2020efficient}. In the latter case, it forms the basis for the  recently proposed ECHO descriptor \cite{mitchel2020echo}, which has been shown to significantly outperform SHOT \cite{tombari2010unique} and other hand-crafted descriptors in terms of overall descriptiveness and robustness to a variety of nuisance factors.

In particular, field convolutions combine extended convolution on surfaces with parallel transport, resulting in a mapping between vector fields that only depends on the Riemannian metric.  Given a feature vector field $X$ and a filter $f$, the construction of the field convolution is straight-forward: At each point $p\in M$ on the surface, values of $X$ in the surrounding neighborhood $q\in\N_p$ are weighted relative to the position of $p$ in the frame determined by $X(q)$, transported to $p$, and aggregated. This approach is agnostic to connectivity and is robust, as the assignment of weights with respect to multiple coordinate systems makes it naturally insensitive to noise and other nuisance factors.  Most importantly, no constraints are placed on filters.

Field convolutions facilitate the construction of highly discriminating yet simple networks, without the need for pooling, normalization, or specialized architecture. The principal module in applications is the field convolution ResNet (\textbf{FCResNet}) block, consisting of two successive field convolutions with a residual connection between the input and output layers \cite{he2016deep}. FCResNet blocks are self-contained and flexible, and can easily be incorporated into isometry-invariant surface learning regimes. In addition, we leverage the connection between field convolutions and the recently proposed state-of-the-art ECHO surface descriptor \cite{mitchel2020echo} to construct a novel final layer specifically designed for labeling tasks with isometry-invariant surface networks, which we refer to as an ECHO block. This block takes vector field channels as input, mapping them to scalar ECHO descriptors which are then fed through an MLP to make predictions, essentially converting the problem to one of image classification in the final layer of the network.

\section{Field Convolution}
\label{s:convolution}
Following the approach of Knoppel {\em et al.}~\cite{Knoppel:2013:GOD}, we represent tangent vectors as complex numbers. Given a surface $M$, at any point $p \in M$  we can assign to the tangent space $T_pM$ an orthonormal basis $\{ \vn{e}_1, \, \vn{e}_2 \}_{p}$. Then $T_{p}M$ can be associated with $\mathbb{C}$ such that for any $\vn{v} \in T_{p}M,$ we have $\vn{v} \equiv r \, e^{i\theta}$, 
with $r = |\vn{v}|$ and $\theta$ the
angle between $\vn{v}$ and $\vn{e}_1$.

Letting $\Gamma(TM)$ be the space of vector fields on $M$, we express the evaluation of a vector field $X \in \Gamma(TM)$ 
at a point $p \in M$, in terms of the frame $\{ \vn{e}_1, \, \vn{e}_2 \}_{p}$, as
\begin{align}
    \begin{aligned}
    X(p) \equiv \rho_{p} \, e^{i \phi_p}.
    \end{aligned} \label{X_notation}
\end{align}
Similarly, for two points $p, \, q \in M$ we denote the logarithm of $p$ with respect to $q$, giving the ``position'' of $p$ in $T_qM$, as
\begin{align}
\log_{q} p \equiv r_{qp} \, e^{i \theta_{qp}}.\label{log_map}
\end{align} 
We denote by $\varphi_{pq}$ the change in angle resulting from the parallel transport $\PT{p}{q}\!: T_qM \rightarrow T_pM$ along the shortest geodesic from $q$ to $p$, such that for any $\vn{v} \in T_qM$, 
\begin{align}
\PT{p}{q}(\vn{v}) \equiv e^{i \varphi_{pq}} \, \vn{v}. \label{transport}
\end{align}
We consider filters belonging to the space of square integrable functions on the complex plane, and define the \textit{field convolution} of a vector field $X \in \Gamma(TM)$ with a filter $f \in L^2(\mathbb{C})$  to be the vector field in $\Gamma(TM)$ with
\begin{align}
   \big(X * f\big)(p) & = \int_M \rho_q \, e^{i (\phi_q + \varphi_{pq})} \, f\left( r_{qp} \, e^{i\left( \theta_{qp} - \phi_{q}\right)} \right) \ dq.  \label{field_conv}
\end{align}
The first term is the parallel transport of the tangent vector $X(q)$ to $T_pM$ and the second term is the evaluation of the filter at the coordinates of $\log_qp$, expressed relative to the frame 
$\left\{ X(q) / \|X(q)\|, X^\perp(q) / \|X(q)\|\right\}.$\footnote{Although the frame is undefined when $\rho_q=0$, the integral remains well-defined as the value of the filter is multiplied by $\rho_q$ in the integrand.}

In practice, we use filters compactly supported within a radius of $\epsilon$ and limit the domain of integration to the geodesic $\epsilon$-ball about $p$. 

Finally, noting that isometries preserve areas, and commute with the action of parallel transport  and the logarithm \cite{gallier2020differential}, it follows that if $\Psi:M\rightarrow N$ is an isometry, we have
\begin{align}
d\Psi\left[\big(X*f\big)(p)\right] = \left[d\Psi(X) * f\right]\big(\Psi(p)\big). 
\end{align}
Or in other words, field convolutions commute with the action of isometries. A detailed proof of this claim can be found in Supplement~\ref{iso_comm}.

\paragraph*{Discretization}
In practice, we discretize a surface $M$ by a triangle mesh with vertices $V$. To every $p \in V$, we associate the collection of vertices $\N_p \subset V$ belonging to the geodesic $\epsilon$-ball about $p$. At each point, real-valued filters $f \in L^2(\mathbb{C})$ are supported on $\log_p\left(\N_p\right) \subset \mathbb{C}$, and parameterized as sums of angular frequencies with band-limit $B$. That is, for any $z = r e^{i \theta} \in \mathbb{C}$ with $\lvert r \rvert \leq \epsilon,$ the evaluation of $f$ at $z$ is expressed as 
\begin{align}
    f(z) = \sum_{m = -B}^{B} f_m(r)\cdot e^{im\theta} \label{f_coeff}
\end{align}
where $f_m(r)\in{\mathbb C}$ is the $m$-th Fourier coefficient of $f$, restricted to radius $r$ and $f_{-m}(r)=\overline{f_m(r)}$ because $f$ is real-valued. We discretize the function $f_m(r)$ using linear interpolation, setting $f_m(r) = \vn{r}^\top\!(r)\, \vn{f}_m$, where $\vn{r}(r)\in{\mathbb R}^N$ is the vector of linear interpolation weights (with $\vn{r}_i(r)\neq0$ only if $i\in\{\lfloor rN/\epsilon \rfloor,\lceil rN/\epsilon\rceil\}$) and $\vn{f}_m\in{\mathbb C}^N$ is the vector of Fourier coefficients at the discrete radii.

Then, letting $\{w_p\} \subset \mathbb{R}_{>0}$ denote the area weights associated with vertices $p \in V$, choosing an arbitrary edge at every vertex to define a frame, and letting $X\in{\mathbb C}^{|V|}$ be a discrete vector field, the evaluation of the field convolution $X * f$ as in Equation~(\ref{field_conv}) at a vertex $p \in V$ is given by
\begin{align}
    \big(X * f\big)(p) =\!\!\! \sum_{\substack{q \in \N_p \\ \lvert m \rvert \leq B}}\!\! w_q \, \rho_q \, e^{i(\phi_q +\varphi_{pq})} \, f_m(r_{qp}) \, e^{im(\theta_{qp} - \phi_q)}. \label{fc_disc}
\end{align}
The values of $w_q, \, \varphi_{pq}, \, r_{qp},$ and $\theta_{qp}$, corresponding to the weight, transport change of angle, geodesic distance, and logarithm for each $p \in V$ and $q \in \N_p$ can be precomputed to speed up training. Similar to \cite{wiersma2020cnns}, we apply rotational offsets $e^{i\beta_{\lvert m \rvert}}$ to the coefficients corresponding to each frequency, providing additional learned degrees of freedom.

\section{Surface CNNs with Field Convolutions}
\label{s:network}
Field convolutions are the principle contribution of this work as they provide a robust and descriptive framework for the spatial propagation of information on surfaces. The goal of this section is to introduce the fundamental building blocks for incorporating field convolutions into isometry-invariant surface learning paradigms. 

\paragraph*{FCResNet Blocks} 
\begin{figure}[!t] 
\centering
\includegraphics[width=\columnwidth]{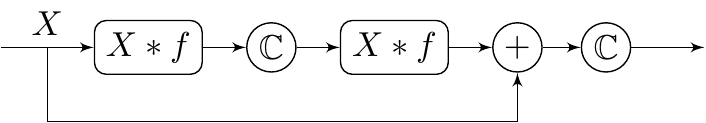}
\caption{The FCResNet block. Here $\mathbb{C}$ denotes the complex ReLU in Equation~(\ref{relu}). \label{fcrn}}
\end{figure}

The atomic unit for field convolutions in surface CNN frameworks is the FCResNet block, which consists of two field convolutions each followed by a non-linearity and a residual connection between the input and output streams (Figure~\ref{fcrn}). They are entirely self-contained, and map vector field features to vector field features without relying on any supporting or complementary convolution operations that are a common fixture in other equivariant approaches \cite{poulenard2018multi, wiersma2020cnns}. As such, they represent a flexible and descriptive layer that can be easily employed in isometry-invariant learning pipelines. 

\paragraph*{ECHO Blocks}
A secondary contribution of this work is the concept of an ECHO block for label-prediction tasks, which leverages the connection between vector fields and the recently proposed ECHO surface descriptor \cite{mitchel2020echo}. Given a scalar signal and a frame field, ECHO descriptors provide an intrinsic, isometry-invariant characterization of the local surface about a feature point in terms of the filter maximizing the response to extended convolution. This filter is constructed by having neighbors of the feature point ``cast a vote'', weighted by the value of the signal at that point, into the filter position corresponding to the position of the feature point, as seen from the neighbor's frame.  A vector field can be used to compute ECHO descriptors at every point $p\in M$, using the magnitude and direction of each vector to define the values of the signal and transformation field at $p$.  

The idea behind ECHO blocks is to convert feature vector fields to pointwise descriptors, turning the task of vector field classification into one of image classification in the final layer of the network. These blocks consist of two steps: 1)~A field convolution layer is used to map the input feature channels to $D$ output feature channels (with $D$ the desired number of descriptors). These are then used to compute pointwise ECHO descriptors, resulting in $H$ isometry-invariant scalar features per channel, where $H$ is the number of samples used to represent the ECHO descriptor. 2)~The $D\times H$ values are linearized and fed to a three-layer MLP. Like field convolutions, the computation of ECHO descriptors relies only on the logarithm map, parallel transport, and the integration weights associated with each vertex. No additional pre-processing is required.

\paragraph*{Linearities and Non-Linearities}
Since we represent tangent vector features as complex numbers, we apply linearities in the form of multiplication by complex matrices in the same manner as is done for real-valued features. However, our linearities do not include translational offsets to preserve commutativity with the action of isometries.

For similar reasons, non-linearities are applied only to the radial components of features as is done in \cite{wiersma2020cnns}. Namely, given a feature vector field $X\in{\mathbb C}^{|V|}$  we apply pointwise ReLUs with a learned offset $b$ such that
\begin{align}
   \textrm{ReLU}_{b}\big(X(p)\big) = \textrm{ReLU}\left(\rho_p + b\right) \, e^{i\phi_p}. \label{relu}
\end{align}

\paragraph*{FCNet: A Generic Surface CNN for Vector Fields}
In our experiments we use a simple, generalizable architecture we call an \textbf{FCNet}, which is simply a series of FCResNet blocks. For labeling tasks, we append an ECHO block to the end of the network to make predictions. For FCNets consisting of three or more layers, we add additional residual connections after every two FCResNet blocks as we find this significantly accelerates training. In all experiments, we take the raw 3D positions of points as inputs and use a learnable gradient-like operation (Supplement~\ref{learned_gradients}) to map them to vector fields which are then fed to the network.
We could also use the intrinsic Heat Kernel Signature~\cite{sun2009concise} as input, thereby obtaining a fully isometry-invariant pipeline. However, as demonstrated by Sharp~{\em et al.}~\cite{sharp2020diffusion}, the 3D coordinates work as well in practice and are easier to compute.

Despite this elementary construction, we show that FCNets achieve state-of-the-art results in a variety of fundamental geometry processing tasks.

\section{Evaluation}
\label{s:evaluation}
We compare our method against leading surface learning paradigms on four benchmarks corresponding to fundamental tasks in geometry processing: classification, segmentation, correspondence, and feature matching.

\subsection{Implementation}
Our framework is implemented using PyTorch Geometric \cite{FeyLenssen2019}. We employ the same, simple FCNet architecture discussed in Section~\ref{s:network} in all of our experiments, varying the number of FCResNet blocks based on task complexity. For label-prediction tasks on large datasets, we append an ECHO block to the end of the network to make predictions. Otherwise we use the magnitudes of the output feature vectors.

As input, we take the 3D coordinates, which are lifted to $16$ tangent vector features in the initial gradient layer, followed by either $32$ or $48$ features in the FCResNet stream. We use the ADAM optimizer \cite{Kingma2014AdamAM} to a cross-entropy loss with an initial learning rate of $0.01$ and a batch size of $1$. We randomly rotate all inputs to ensure there are no consistencies in the spatial embedding of shapes. 

Our pre-processing regime parallels \cite{wiersma2020cnns}, omitting the operations necessary to support their multi-scale and pooling operations.  All shapes are normalized to have unit surface area and we use the Vector Heat Method \cite{Sharp:2019:VHM} to compute the geodesic $\epsilon$-ball $\N_{p} \subset V$ corresponding to each vertex $p \in V$, in addition to the logarithm and parallel transport associated with each edge $(p, \, q) \in \{p\} \times \N_{p}$. Area weights are assigned in the standard way, using one third of the vertex's one-ring area, and are normalized by the sum of the weights within the geodesic $\epsilon$-ball. While we process shapes as triangle meshes in our experiments, we note that recent work by Sharp \etal \cite{Sharp:2020:LNT} has made possible efficient computations of logarithmic parameterizations and vector transport on point clouds, with which our method can be extended to analyze point cloud shape data.

\subsection{Classification}

\begin{table}[t]{
    
    \centering
    \begin{tabular}{ll}
    Method              & Accuracy \\ \hline \hline
    \multicolumn{1}{l|}{FC (ours)}         & \textbf{99.2\%}   \\
    \hline
    \multicolumn{1}{l|}{DiffusionNet \cite{sharp2020diffusion}} & 98.9\% \\
    \multicolumn{1}{l|}{MeshWalker \cite{lahav2020meshwalker}} & 97.1\% \\
    \multicolumn{1}{l|}{HSN \cite{wiersma2020cnns}} & 96.1\% \\
    \multicolumn{1}{l|}{MeshCNN \cite{hanocka2019meshcnn}}     & 91.0\%  \\
    \multicolumn{1}{l|}{GWCNN \cite{gwcnnEzuz}}        & 90.3\%  \\
    \end{tabular}
    \caption{Classification accuracy on the SHREC '11 Dataset \cite{lian2011} \label{class_results}. }
    } 
\end{table}

\begin{table}[t]{
    \centering
    \begin{tabular}{lll}
    Method                           & \# Features             & Accuracy \\ \hline \hline
    \multicolumn{1}{l|}{FC (ours)}         & \multicolumn{1}{l|}{3}  & \textbf{92.9\%}  \\
    \hline
    \multicolumn{1}{l|}{MeshWalker \cite{lahav2020meshwalker}}     & \multicolumn{1}{l|}{NA}  & 92.7\%  \\
    \multicolumn{1}{l|}{MeshCNN \cite{hanocka2019meshcnn}}     & \multicolumn{1}{l|}{5}  & 92.3\%  \\
    \multicolumn{1}{l|}{DiffusionNet \cite{sharp2020diffusion}} & \multicolumn{1}{l|}{16} & 91.5\% \\
    \multicolumn{1}{l|}{HSN \cite{wiersma2020cnns}}         & \multicolumn{1}{l|}{3}  & 91.1\% \\
    \multicolumn{1}{l|}{SNGC \cite{haim2019surface}}         & \multicolumn{1}{l|}{3}  & 91.0\%  \\
    \multicolumn{1}{l|}{PointNet++ \cite{qi2017pointnet}}  & \multicolumn{1}{l|}{3}  & 90.8\%  \\
    \end{tabular}
    \caption{Segmentation accuracy on the composite dataset of \cite{maron2017convolutional}. \label{segmentaion_results}}
    }
\end{table}

First, we use an FCNet with two FCResNet blocks to classify meshes in the SHREC '11 dataset \cite{lian2011}, containing $30$ shape categories. Filters are supported on geodesic neighborhoods of radius $\epsilon = 0.2$ and are parameterized using $N = 6$ radial samples with band-limit $B = 2$.  Due to the small scale of the task, we omit the ECHO block in the final layer and instead use a global mean pool over the feature magnitudes to give a prediction.  As in prior works \cite{hanocka2019meshcnn, wiersma2020cnns, sharp2020diffusion}, we train on 10 samples per class and report results over three random samplings of the training data. Our FCNet converges quickly, and we train on just 30 epochs -- far fewer than the 100 or more used in previous work.

Results are shown in Table~\ref{class_results}. Due to the wide adoption of the dataset, we only list the results of methods achieving a classification accuracy of $90\%$ or higher. Our simple FCNet achieves the highest reported accuracy, reaching a classification rate of $100\%$ on two of the three random samplings of the training data. Like HSN and DiffusionNet who also report high classification accuracy, our FCNet uses relatively few parameters compared to other networks and is agnostic to both mesh connectivity and isometric deformations -- all providing a significant advantage on the SHREC '11 dataset which has a small number of training samples and consists of poor-quality meshes with in-class deformations mainly limited to rigid articulations. The superior performance of our FCNet is likely due to the descriptiveness of field convolutions, as HSN uses specially parameterized  filters.

\subsection{Segmentation}

\begin{figure}[!t] 
\centering
\includegraphics[width=\columnwidth]{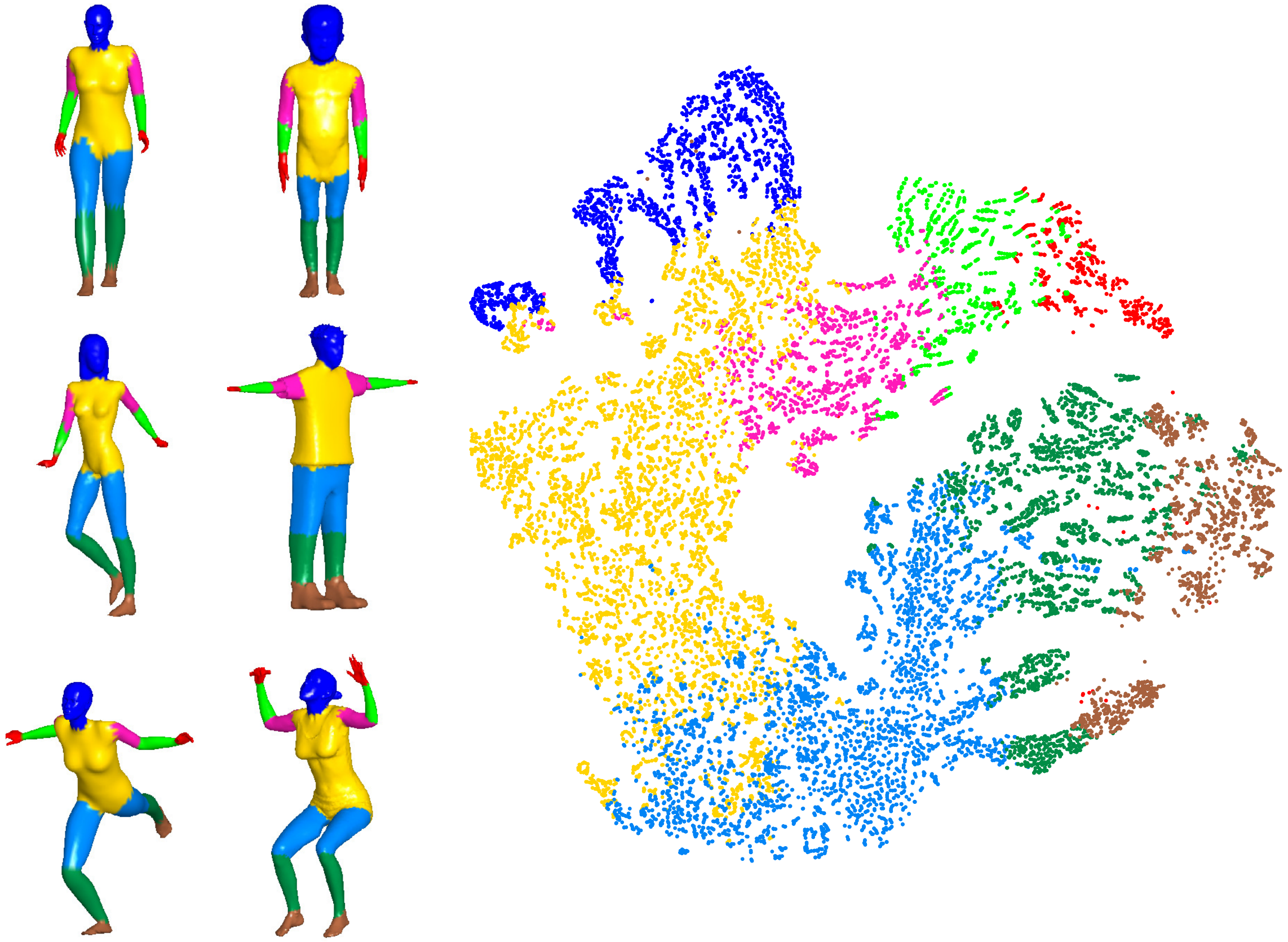}
\caption{We visualize the descriptors computed in our FCNet's ECHO block in the segmentation task.  \label{seg_viz} Left: Models from the test split of the composite dataset \cite{maron2017convolutional}, color-coded by ground-truth labels. Right: 2D projections of the descriptors using t-SNE \cite{van2008tSNE}.
}
\end{figure}
Next, we apply our field convolution framework to the task of human body segmentation,  using the dataset proposed by \cite{maron2017convolutional}, which consists of a composite of various human shape datasets \cite{adobe_2016, SCAPE, giorgi2007shape, MIT, Bogo:CVPR:2014}. The varied nature of the collection of models in terms of human subjects, acquisition method, and connectivity serve to test both descriptiveness and robustness to variety of nuisance factors. 

We use an FCNet with four successive FCResNet blocks $(N = 6, \, B = 2, \, \epsilon = 0.2)$ followed by an ECHO block, trained to predict a body part annotation for each point on the mesh. The ECHO block computes $D = 32$ descriptors with $H = 33$ samples (corresponding to three samples per geodesic radius) for a total of $1056$ scalar feature channels. The three-layer MLP first maps these features to $256$ channels, then $124$, and finally to the desired number of output channels.  Due to the large number of vertices per model, we downsample each mesh to $1024$ vertices using farthest point sampling, an approach also used by \cite{hanocka2019meshcnn, wiersma2020cnns}. Our network converges quickly and we train for only 15 epochs with a label smoothing regularization \cite{szegedy2016rethinking} factor of $0.2$.

Results in the form of the percentage of correctly classified vertices across all test shapes are shown in Table~\ref{segmentaion_results}. As in the classification experiments, we only list the results of methods that achieve a segmentation accuracy of $90\%$ or higher on the dataset. Again, our basic network achieves state-of-the-art results, outperforming all other methods with a minimal number of input features. The improvement due to field convolutions is especially evident relative to other techniques that employ surface convolutions, such as HSN~\cite{wiersma2020cnns} and DiffusionNet~\cite{sharp2020diffusion} approaches.

To understand the features learned by our network, we use t-SNE\cite{van2008tSNE, scikit-learn} to visualize the descriptors computed in the ECHO block for all models in the test dataset, color-coded using the ground-truth labels (Figure~\ref{seg_viz}).  We observe a distinct clustering of points,  not only corresponding to similarly labeled regions but also reflecting the connectivity between adjacent regions on the meshes. This suggests that our FCNet is able to learn at least some measure of intrinsic similarities between shapes, despite starting with the extrinsic 3D coordinates as input.

\subsection{Correspondence} \label{dense_corr}

\begin{figure}[!t] 
\centering
\includegraphics[width=\columnwidth]{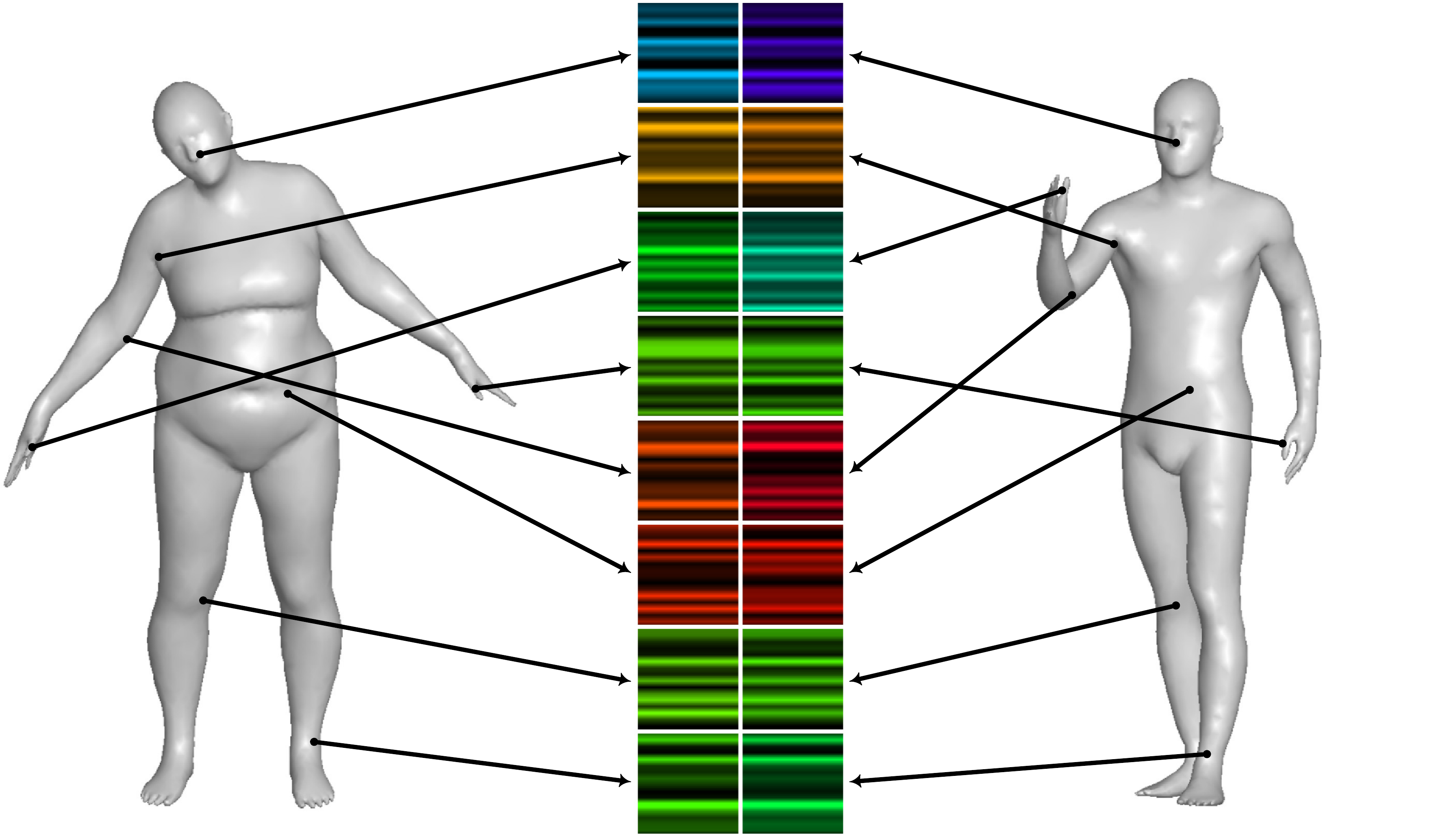}
\caption{FCNet features at corresponding points on models in the remeshed FAUST dataset \cite{donati2020deep}. Features are drawn using the HSV scale -- hue encodes the absolute magnitude and value encodes the relative magnitude with  saturation fixed at one. \label{faust_viz} }
\end{figure}

\begin{figure}[!t] 
\centering
\includegraphics[width=\columnwidth]{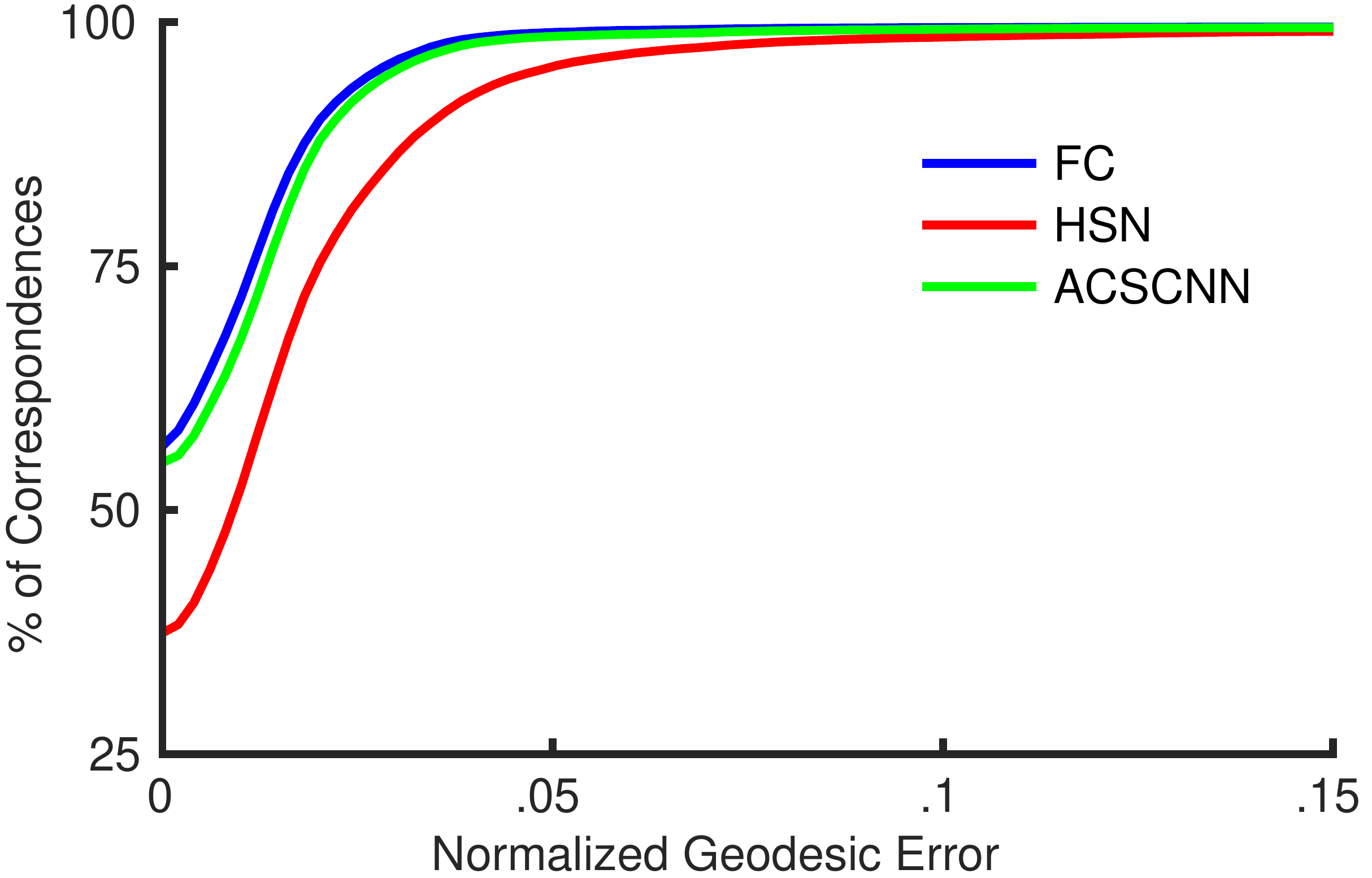}
\caption{Percentage of correspondences for a given geodesic error on the remeshed FAUST dataset using field convolutions (FC), HSN, and ACSCNN. \label{corr_results} }
\end{figure}

\begin{figure}[!t] 
\centering
\includegraphics[width=\columnwidth]{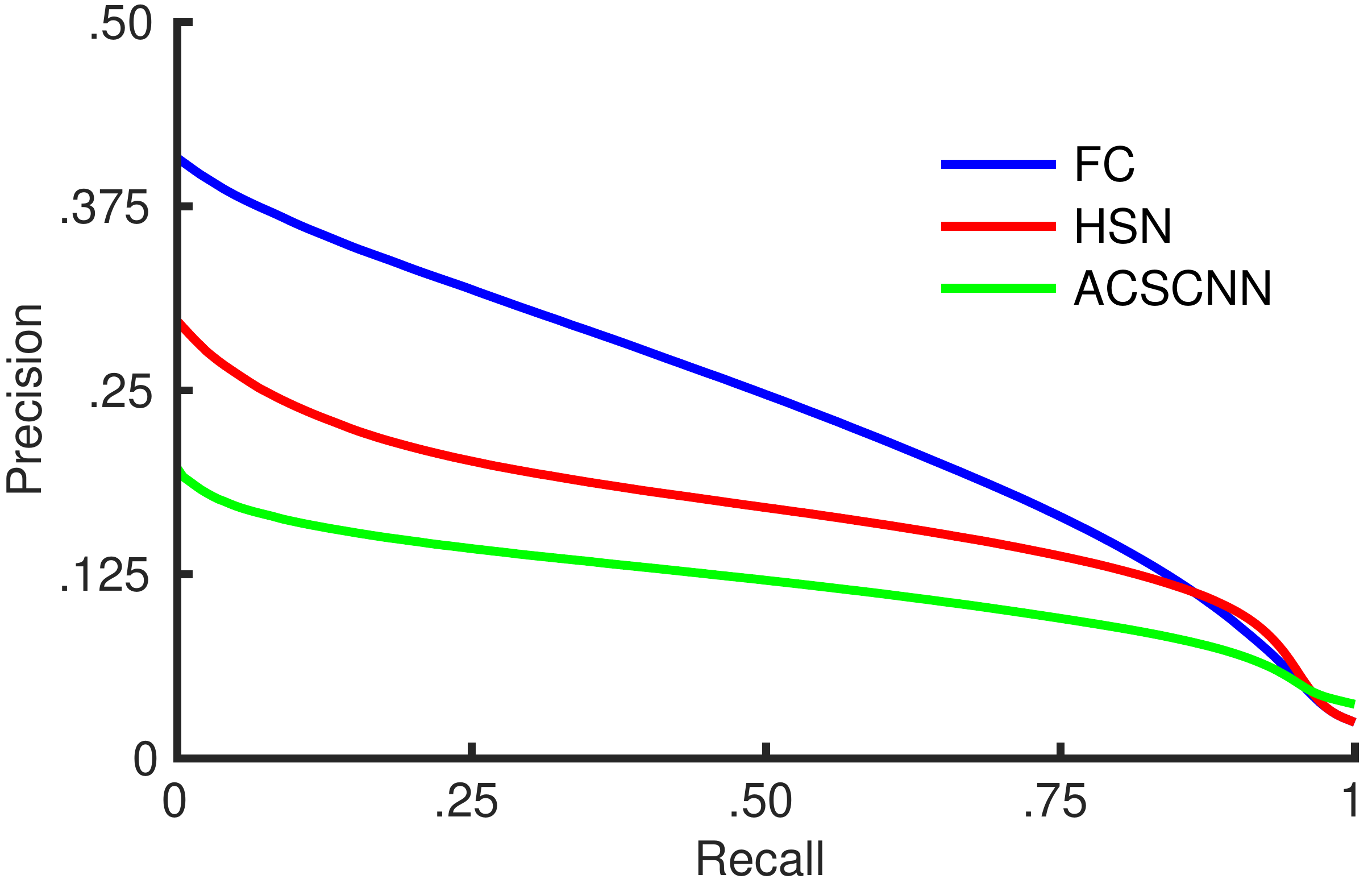}
\caption{Results of feature matching evaluations on the SHREC 2019 Isometric and Non-Isometric Shape Correspondence dataset \cite{Dyke:2019:track} in the form of the mean precision-recall curves. \label{matching_results}}
\end{figure}

Here we use an FCNet to find pointwise correspondences between similar shapes. Over the last half-decade, the FAUST dataset \cite{Bogo:CVPR:2014} has become the \textit{de facto} standard for evaluating network performance in correspondence tasks and many recent approaches have achieved near-perfect accuracy on the dataset \cite{fey2018splinecnn, de2020gauge, li2020shape}. However, shapes in the dataset share the same connectivity, and there has been some question as to whether these methods have primarily learned the mesh graph structure, rather than deformation-invariant characterizations of the shape themselves \cite{sharp2020diffusion}. To this point, we perform evaluations on a fully remeshed version of the dataset \cite{donati2020deep}, a more challenging task better representative of real-world applications. As in prior work, we train on the first $80$ models out of the $100$ in the dataset, and use the remainder for testing.

We train an FCNet to predict the indices of corresponding vertices on a template shape. Due to the degree of precision required by this task, we use a deeper network, consisting of eight FCResNet blocks $(N = 3, \, B = 1, \, \epsilon = 0.05)$ followed by an ECHO block  $(D = 12, \, H = 13)$  with a $124$--$64$--$32$ MLP, and additional residual connections after every two FCResNet blocks.  To make predictions, we add two linear layers after the ECHO block, taking the $32$ features first to $256$ channels, and then to the number of vertices on the template shape, with a $p = 0.5$ dropout layer in-between. Visualizations of some of the $32$ channel features in our FCNet at the bottleneck before the dense final layers are shown in Figure~\ref{faust_viz}.

Prior methods have typically used high-dimensional SHOT \cite{tombari2010unique} descriptors as inputs for this task, which we feel to be unnecessary due to the expressiveness of the field convolution framework. As such, we train HSN and ACSCNN \cite{li2020shape} with raw 3D coordinates inputs for comparison -- two recent methods which have reported state-of-the-art results in similar classification tasks. The results are shown in Figure~\ref{corr_results}, giving the percentage of total correspondences as a function of the normalized geodesic error. Our FCNet achieves the best performance, followed by ACSCNN. 

Recent spectral-based networks, ACSCNN and DiffusionNet \cite{sharp2020diffusion}, have significantly outperformed comparable equivariant networks in correspondence related tasks. This is likely for two reasons: 1) In contrast to the local patch-based convolution operators used in equivariant networks, spectral-based convolutions are formulated in a Laplace-Beltrami basis, providing an inherently global characterization of shape less sensitive to point-wise noise or local mislabeling;  2) The ability to essentially band-limit convolutions by working in basis of low-frequency eigenfunctions allows spectral-based networks to easily scale to high resolutions whereas equivariant networks must decrease both filter support and the number of parameters to process the same meshes. This makes the performance of our FCNet particularly notable as it suggests that the network is able to overcome the relative limitations of equivariant frameworks in dense correspondence tasks specifically due to the robust construction of field convolutions as a scattering operation -- keeping them stable despite smaller supports -- and due to their descriptiveness, the latter of which does not diminish significantly even with fewer parameters.

\subsection{Feature Matching}
\label{matching}

\begin{figure}[!t] 
\centering
\includegraphics[width=\columnwidth]{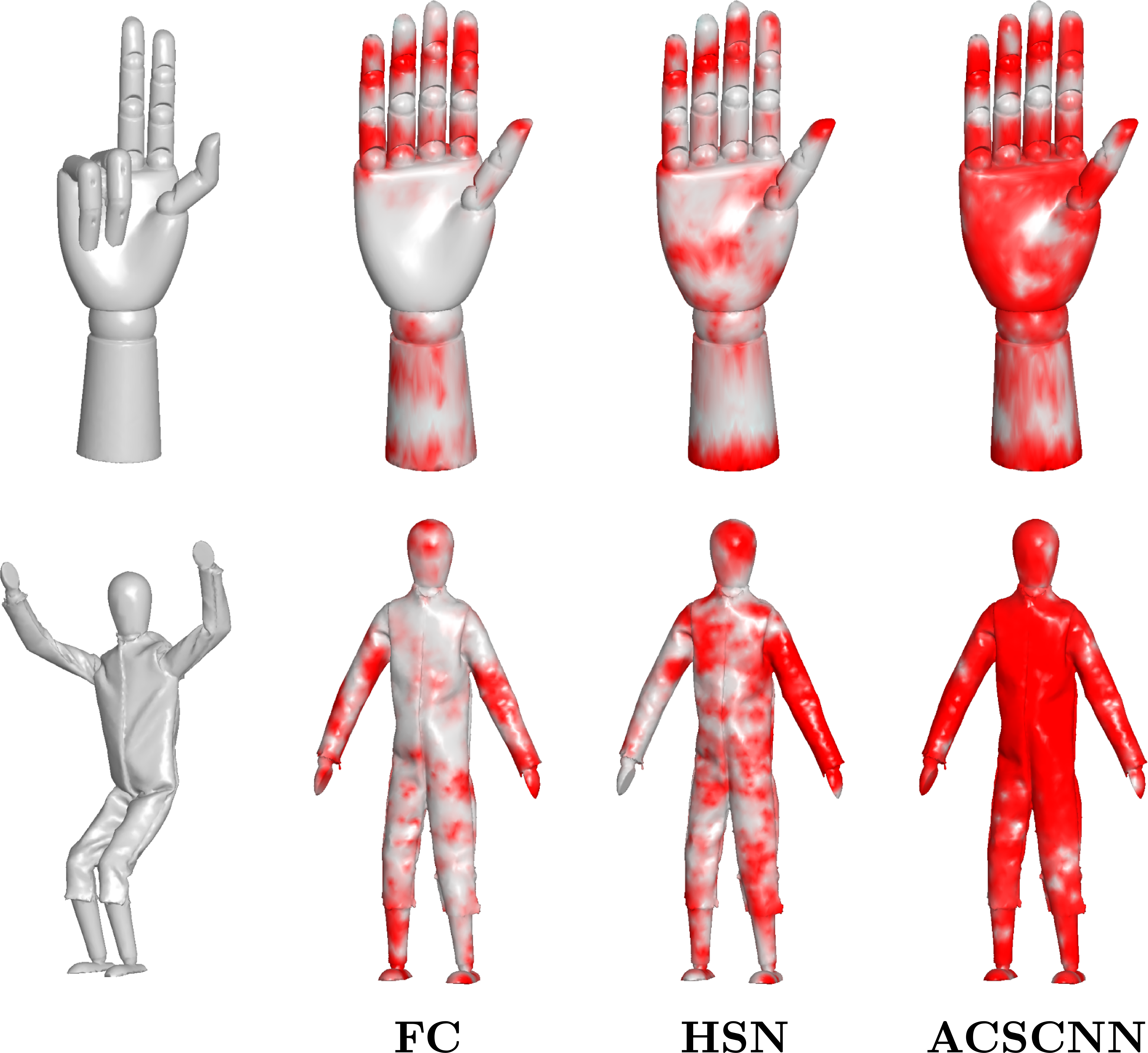}
\caption{%
Feature-space distances: For each feature from the models on the right, we rank order the features on the model to the left by feature distance. Vertices are then colored from gray to red, showing how deeply one must traverse the rank ordered list before encountering a corresponding feature.
\label{feat_compare}
}
\end{figure}

Last, we train an FCNet to compute point-wise surface feature descriptors on shapes from the SHREC 2019 Isometric and Non-Isometric Shape Correspondence dataset \cite{Dyke:2019:track}. The dataset consists of $50$ meshes constructed from 3D scans of a jacketed humanoid figurine and a bare and gloved articulated wooden hand with $76$ pre-defined pairs of meshes.  We consider this dataset to be extremely challenging with significant non-isometric deformations and topological changes between pairs; as real-world scans the meshes also contain noise, varying triangulations, occluded geometry and various other sources of interference. 

To ensure an even distribution of meshes in both the training and testing data, we group all pairs into three categories based on scan source (humanoid, hand, and gloved hand) and randomly select $20\%$ of the pairs in each category to form the test split. We randomly sample $2048$ points on both meshes in each pair and use the ground truth correspondence to assign corresponding and non-corresponding points. We learn compact, $16$-dimensional descriptors at each point using a twin network \cite{litman2013learning, masci2015geodesic, sun2020embedded}, where each mesh in a pair is processed by the same network and a twin loss function is minimized, weighting the descriptor distances between corresponding and non-corresponding points. We use precision-recall curves to evaluate performance on the test pairs, as they have been shown to well-characterize feature descriptiveness  \cite{ke2004pcasift, mikolajczyk2005GLOH} and are the standard metric in the surface feature descriptor literature \cite{tombari2010unique, salti2014shot, guo2013rotational, guo2016comprehensive, mitchel2020echo}. A detailed explanation of our experimental regime can be found Section~\ref{feature_exp} of the supplement. 

We train an FCNet consisting of eight FCResNet blocks $(N = 6, \, B = 1, \, \epsilon = 0.1)$  on the downsampled $2048$-vertex mesh pairs, using the magnitudes of the output features as point-wise descriptors. HSN and ACSCNN are trained on the downsampled and full-resolution meshes, respectively. We report results averaged over three random samplings of the test-train split (Figure~\ref{matching_results}); to ensure fair comparisons, we compute the average precision-recall curves over all test pairs using the same set of correspondences for all methods. Our FCNet achieves the best performance by a significant margin, followed by HSN. The difference is likely explained by the increased descriptiveness of field convolution and its robust formulation as a scattering operation, making it better able to characterize flat, featureless areas (Figure~\ref{feat_compare}, palm of the hand) and insensitive to high-frequency perturbations of the surface (Figure~\ref{feat_compare}, folds in the figurine jacket), as compared to the gathering-based convolution operations used by HSN which rely on strongly constrained filters.  We believe that ACSCNN under-performs relative to the other methods because methods like  ACSCNN which depend on the (global) spectral decomposition of the Laplace-Beltrami operator are less stable in the presence of non-isometric deformations, geometric occlusions, and changes in topology between corresponding pairs. While still not giving excellent performance, methods like FCNet and HSN, which use filters with local support, tend to be more robust.

\begin{comment}
We train an FCNet consisting of eight FCResNet blocks $(N = 6, \, B = 1, \, \epsilon = 0.1)$  on the downsampled $2048$-vertex mesh pairs, using the magnitudes of the output features as point-wise descriptors. HSN and ACSCNN are trained on the downsampled and full-resolution meshes, respectively. We report results averaged over three random samplings of the test-train split (Figure~\ref{matching_results}); to ensure fair comparisons, we compute the average precision-recall curves over all test pairs using the same set of correspondences for all methods. Our FCNet achieves the best performance by a significant margin, followed by HSN. The difference in performance is likely explained by the robust formulation of field convolution as a scattering operation, making it less sensitive to the various nuisance factors present in the dataset, as compared to the gathering-based convolution operations used by HSN.  ACSCNN significantly under-performs relative to the other methods. We believe this is because methods like  ACSCNN which depend on the (global) spectral decomposition of the Laplace-Beltrami operator are unstable in the presence of complex (non-isometric) deformations, geometric occlusions, and changes in topology between corresponding pairs. While still not giving excellent performance, methods like FCNet and HSN, which use filters with local support, tend to be more robust.
\end{comment}

\subsection{Performance}
Field convolutions are among the most efficient equivariant convolution operations, requiring few parameters per convolution operation. On an RTX $2080$ GPU and $3.8$ GHz CPU, our deepest FCNet trains at approximately $3$  min/epoch on the full resolution meshes in the dense correspondence task. Field convolutions use a similar number of parameters as HSN \cite{wiersma2020cnns} per convolution and with half the memory footprint. HSN's multi-stream convolutions learn a weight matrix for the radial profile and rotational offset corresponding to each stream and the connections between them, resulting in $(N + 1) \, S^2$ total parameters per convolution, with $N$ the number of radial samples and $S$ the number of streams. Similarly, we learn a complex radial profile and rotational offset for each non-negative frequency up to the number of band-limited frequencies with  $ N (2 B + 1) + B + 1$ total parameters per convolution. In the classification and segmentation experiments, HSN reports results using $N = 6$ radial samples and $M = 2$ streams resulting in $28$ total parameters per convolution. In the same experiments, our FCNet achieves state-of-the-art performance with $33$ total parameters per convolution, as we use filters with band-limit $B = 2$ and the same number of radial bins. However, HSN stores features for both streams, increasing spatial complexity by a factor of two.

More generally, we see our state-of-the-art results on the segmentation task using the composite dataset \cite{maron2017convolutional} as particularly notable in that other top-performing methods, including MeshCNN \cite{hanocka2019meshcnn} and HSN, use the deepest versions of their network for this task despite the small number of labels involved (eight classes), presumably because of the large size of the training dataset.  Our FCNet outperforms these networks with a much shallower architecture and only in the dense correspondence and feature matching tasks -- both of which involve learning granular distinctions between large numbers of similar points -- do we increase the depth of our network. This suggests that unlike most networks, the depth of an FCNet (or other network built on field convolutions) necessary to achieve good performance is not strongly dependent on the size of the dataset, and scales primarily with task complexity.

\section{Conclusion}
\label{s:conclusion}
We present a novel definition of surface convolution acting on vector fields,  combining invariant spatial weighting with the parallel transport of features in a scattering operation \textit{while placing no constraints on the filters themselves}. This formulation is highly descriptive, insensitive to a variety of nuisance factors, and straight-forward to implement; with it, we construct simple networks that achieve state-of-the-art results in fundamental geometry-processing tasks. 

While the complexity of our method is comparable to existing equivariant approaches, it shares the same drawbacks as filter supports and parameter counts must be limited to process meshes at full resolution. More generally, existing successful surface learning frameworks (including ours) are designed to handle only isometric or nearly-isometric shape deformations and fail to achieve adequate performance in the presence of the kinds of complex deformations, geometric occlusions, and topological changes found in real shape data. In the future, we plan to expand our framework to handle more challenging classes of deformations, beginning with invariance to conformal automorphisms. 

%%%% References
{\small
\bibliographystyle{ieee_fullname}
\bibliography{ref}
}

%% Supplementary Material
\clearpage
\appendix
\section{Supplement}
\label{s:supplement}
\subsection{Field convolutions commute with isometries}
\label{iso_comm}
Here we offer a detailed proof of the claim made in Section 4 that field convolution commutes with the action of isometries. That is, given any $X \in \Gamma(TM)$ and filter $f \in L^2(\mathbb{C})$, if $\Psi:M\rightarrow N$ is an isometry, then
\begin{align}
d\Psi\left[\big(X*f\big)(p)\right] = \left[d\Psi(X) * f\right]\big(\Psi(p)\big). 
\end{align}

To see this, consider surfaces $M$ and $N$ and any two points $p\in M$ and $p' \in N$. Let $\N \subset M$ and $\N' \subset N$ be $\epsilon-$balls about the points and suppose that $\N$ and $\N'$ are isometric. That is, there exists a map $\Psi:M\rightarrow N$ taking $p$ to $p'$ and satisfying $\forall q_i \in \N,$
\begin{align*}
    d\left(q_0, \, q_1\right) = d\left(q_0', \, q_1'\right),  \quad q_i' = \Psi(q_i) \in \N',
\end{align*}
where $d\left( \, \cdot \, , \, \cdot \,\right)$ is the geodesic distance. 

Let $X' \in \Gamma(TN)$ be the
push-forward of $X$ under $d\Psi$
where, using the tangent space representation of Knoppel {\em et al.}~\cite{Knoppel:2013:GOD}, $\eval{X'}{p'} = \rho'_{p'} \, e^{i \phi'_{p'}}$. For any two points $a, b \in M$, denote the logarithm of $a$ with respect to $b$ and the change in angle resulting from the parallel transport along the shortest geodesic from $b$ to $a$ as $\log_{b}a = r_{b a} \, e^{i \theta_{b a}}$ and $\varphi_{a b}$, respectively.  It follows that $\forall q \in \N$ \cite{gallier2020differential},
\begin{gather*}
    \begin{aligned}
    \begin{aligned}
    \rho'_{q'} &= \rho_{q} \\
    r_{q'p'} &= r_{qp} \\
    \end{aligned}
    & 
    \begin{aligned}
    \quad \textrm{and} \quad \\
    \quad \textrm{and} \quad \\
    \end{aligned}
    \begin{aligned}
    \phi'_{q'} &= \phi_{q} + \psi_q, \\
    \theta_{q'p'} &= \theta_{qp} + \psi_q,
    \end{aligned}
    \end{aligned} \\
    \varphi_{p'q'} = \varphi_{pq} + \psi_p - \psi_q,
\end{gather*}
where $\psi_{p}$ is the angle of rotation corresponding to the action of the differential $\eval{d\Psi}{p}$, taking vectors in $T_pM$ to $T_{p'}N.$ (Recall that as $\Psi$ is an isometry, $d\Psi$ is an orthogonal transformation.) Then, in the expression for the field convolution
\begin{align*}
   \eval{\left(X' * f\right)}{p'} & \!=\!\!
   \int_M \rho_{q'}e^{i (\phi_{q'} + \varphi_{p'q'})} \, f\left( r_{q'p'} e^{i\left( \theta_{q'p'} - \phi_{q'}\right)} \right) \ dq'
\end{align*}
we have
\begin{align*}
    \rho'_{q'} \, e^{i\left(\phi'_{q'} + \varphi_{p'q'}\right)} &= \rho_q \, e^{i\left(\phi_q + \varphi_{pq} + \psi_p\right)}, \\
    r_{q'p'} \, e^{i\left(\theta_{q'p'} - \phi'_{q'}\right)} &= r_{qp} \, e^{i\left(\theta_{qp} - \phi_{q}\right)},
\end{align*}
with the measures $dq$ and $dq'$ satisfying $dq' = dq$ since $\eval{d \Psi}{q}$ is an orthogonal transformation. From the definition of field convolution, this gives $\eval{\left(X' * f\right)}{p'} = e^{i\psi_p} \eval{\left(X * f\right)}{p}$  which is equivalent to $d\Psi\left[\big(X*f\big)(p)\right] = \left[d\Psi(X) * f\right]\big(\Psi(p)\big)
$ as desired.

\subsection{Learned Gradients}
\label{learned_gradients}
In practice, inputs to surface CNNs are often scalar features, such as the raw 3D positions of points. To lift such features to a vector field, we use a learnable operation analogous to a weighted gradient calculation.  For any function $\xi \in L^2(V)$ we learn the magnitude and direction of its ``gradient" separately, with respect to compactly supported radially isotropic filters $f_1, \, f_2 \in L^2(\mathbb{C})$. That is, we learn the vector field $\Phi_{f_1}:V \rightarrow {\mathbb C}$ and scalar field $P_{f_2}:V \rightarrow \mathbb{R}$ with
\begin{align}
     \Phi_{f_1}(p) &= e^{i\beta}\sum_{q \in {\mathcal N}_p} w_q \, \left(\xi(q) - \xi(p) \right) \, f_1(r_{pq}) \,  e^{i\theta_{pq}}, \label{grad_dir} \\
     P_{f_2}(p) &= \sum_{q \in {\mathcal N}_p} w_q \, \xi(q) \, f_2(r_{pq}) \label{grad_mag}
\end{align}
with $w_q, \, r_{pq}, \, \theta_{pq}$ defined as in Equation~(7) (the latter two parameters corresponding to $\log_{p}q$) and $\beta$ a learnable rotational offset. Using these, we define the ``gradient'' of $\xi$ with respect to $f_1$ and $f_2$ as the vector field
\begin{align}
    P_{f_2}^2(p)\frac{\Phi_{f_1}(p)}{\left\lVert \Phi_{f_1}(p) \right\rVert}
\end{align}
While this approach ensures that scalar features are passed directly to vector fields,  we do not consider it to be a critical part of our framework and it can be replaced by a linear layer with only a small decrease in performance.

\subsection{Feature Matching Experiments} \label{feature_exp}
Here we provide a detailed explanation of how our feature matching experiments are performed in Section~6.5. Each pair in the SHREC 2019 Correspondence Dataset \cite{Dyke:2019:track} consists of a \textit{model} mesh $V_M$ and a $\textit{scene}$ mesh $V_S$, with the dense ground-truth correspondence mapping the latter to the former. We randomly generate correspondences $C_{SM} = \{ (s_{i}, \, m_{i}) \} \subset V_{S} \times V_{M}$ and non-correspondences $N_{SM} = \left(V_{S} \times V_{M}\right) \setminus C_{SM}$ by selecting $2048$ points on both the model and the scene mesh using farthest point sampling, mapping the sampled scene points to the model mesh using the ground truth correspondence, and associating each mapped scene point to the geodesically nearest sampled point on the model. 

In training, the objective of the network is to make the outputs for corresponding and non-corresponding pairs as similar and dissimilar as possible, respectively \cite{litman2013learning, masci2015geodesic}. To this end we use a twin network, wherein each mesh in a pair is fed to the same network which learns a compact $16$-dimensional descriptor $F$ at each point. Specifically, for each pair in each epoch, we randomly subsample $512$ pairs of corresponding and non-corresponding points, $P_{SM} = C_{SM}^{512} \cup N_{SM}^{512}$ and minimize the twin loss \cite{sun2020embedded}
\begin{gather}
    \begin{aligned}
    L\left(P_{SM}\right) & = \sum_{(s, \, m) \in P_{SM}} \alpha_{s, m}  {\lVert F_S(s) - F_M(m) \ \rVert}^2  + \\
    &\left(1 - \alpha_{s, m}\right) \textrm{max} \left( 0, \, 5 - {\lVert F_S(s) - F_M(m) \ \rVert}^2\right),
    \end{aligned} \label{twin_loss}
\end{gather}
where $\alpha_{s, m} = 1$ if $(s, \, m) \in C_{SM}$ or is set to a random variable between $0$ and $0.2$ otherwise. 

We compute precision-recall curves as follows. Given a sampled point in the scene mesh $s \in V_S$, we sort all sampled model points based on descriptor distance, giving  $\{m_1, \, \ldots , \, m_{K} \} \subset V_M$, with 
\begin{align*}
    \lVert F_S(s) - F_M(m_i) \rVert \leq \lVert F_S(s) - F_M(m_{i+1})\rVert,
\end{align*}
for $ \ 1 \leq i \leq K - 1.$ We define $\mathcal{M}_p \subset V_M$ to be the set of sampled model points that are valid matches with $p$, which consists of all sampled model points whose ground-truth correspondence lies within a geodesic ball of radius $0.05$ about $p$. While this corresponds to a slightly more relaxed definition of correspondence, we find that all methods perform better maintaining a stricter notion of correspondence during training. Then, following \cite{shilane2004princeton, mitchel2020echo} the precision ${\mathcal P}_{p}$ and recall ${\mathcal R}_{p}$ assigned to $p$ are defined as functions of the top $r$ model keypoints,
\begin{align}
   {\mathcal P}_p(r) &= \frac{ \abs{ {\mathcal M}_p \cap \left\{m_{i}\right\}_{i \leq r}}}{r}, \label{precision} \\
    {\mathcal R}_p (r) &= \frac{ \abs{ {\mathcal M}_p \cap \left\{m_{\,i}\right\}_{i \leq r}}}{\abs{{\mathcal M}_p}}.    \label{recall}
\end{align}

\end{document}